\documentclass{article}
\usepackage{spconf}

\usepackage{textcomp}

\usepackage{amsmath}
\usepackage{amssymb}
\usepackage{booktabs}
\usepackage{cite}
\usepackage{color}
\usepackage{enumitem}
\usepackage{gensymb}
\usepackage{graphicx}
\usepackage{latexsym}
\usepackage{lipsum}
\usepackage{multirow}
\usepackage{nonfloat}
\usepackage[all]{nowidow}
\usepackage{pifont}
\usepackage[caption=false,farskip=4pt]{subfig}
\usepackage{url}
\usepackage{xfrac}

\usepackage[pagebackref=false,breaklinks=true,colorlinks=true,bookmarks=false]{hyperref}

\def\sdagger{$^{\dagger}$}
\def\sddagger{$^{\ddagger}$}
\def\q{{\mathbf q}}
\def\x{{\mathbf x}}
\def\t{{\mathbf t}}
\def\r{{\mathbf r}}
\def\y{{\mathbf y}}
\def\k{{\mathbf k}}
\def\L{{\cal L}}

\title{Deep Quantigraphic Image Enhancement via Comparametric Equations}
\name{Xiaomeng Wu\sdagger, Yongqing Sun\sddagger, and Akisato Kimura\sdagger}
\address{\sdagger Communication Science Laboratories, NTT Corporation \\
\sddagger Computer and Data Science Laboratories, NTT Corporation}

\begin{document}
\hyphenation{}
\ninept
\maketitle
\begin{abstract}
Most recent methods of deep image enhancement can be generally classified into two types: decompose-and-enhance and illumination estimation-centric. The former is usually less efficient, and the latter is constrained by a strong assumption regarding image reflectance as the desired enhancement result. To alleviate this constraint while retaining high efficiency, we propose a novel trainable module that diversifies the conversion from the low-light image and illumination map to the enhanced image. It formulates image enhancement as a comparametric equation parameterized by a camera response function and an exposure compensation ratio. By incorporating this module in an illumination estimation-centric DNN, our method improves the flexibility of deep image enhancement, limits the computational burden to illumination estimation, and allows for fully unsupervised learning adaptable to the diverse demands of different tasks.
\end{abstract}

\section{Introduction}
\label{s:introduction}

High-quality input images are critical for many machine vision tasks, such as visual surveillance, autonomous driving, and computational photography~\cite{LiGHJCGL21}. However, images are often captured under unavoidable environmental and technical constraints, such as inadequate or non-uniform lighting and limited exposure times, leaving details hidden in the dark. Therefore, it is necessary to improve the visibility of such low-light images prior to downstream processing and analysis. Recently, deep image enhancement techniques have received a lot of attention and generally fall into two main categories: decompose-and-enhance and illumination estimation-centric.

\begin{figure}[t]
\centering
\subfloat[Decompose-and-enhance methods~\cite{WeiWY018,XuYYL20,Yang0FW020,ZhangGMLZ21}]{
\includegraphics[width=.98\linewidth]{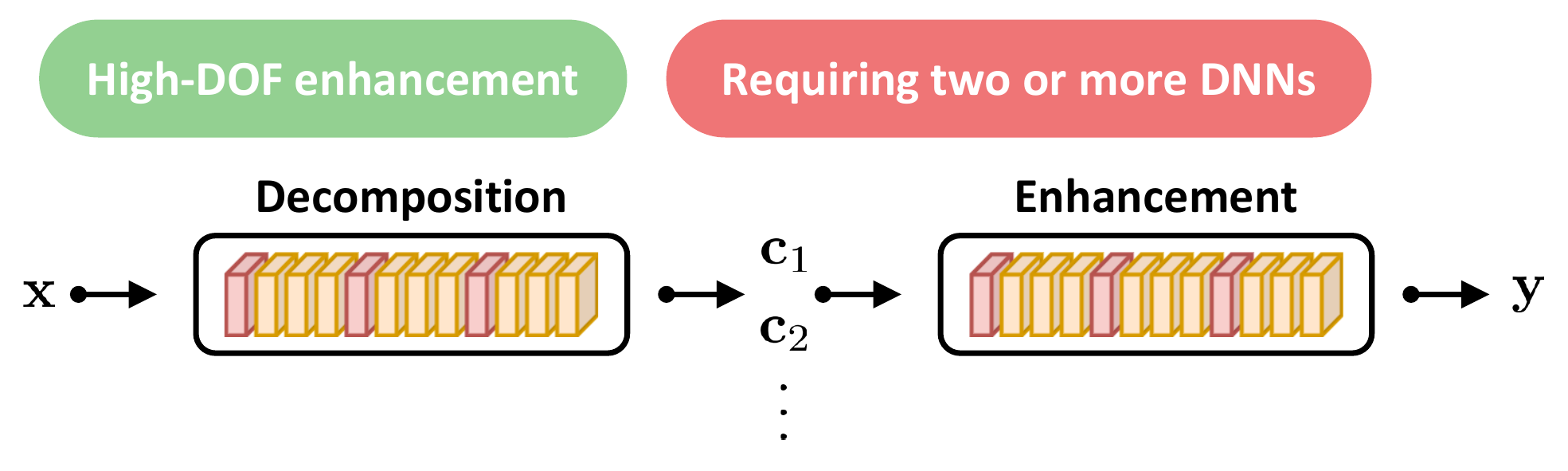}
\label{subf:decompose-and-enhance}}
\\
\subfloat[Illumination estimation-centric methods~\cite{WangZFSZJ19,ZhangDZW20,Liu0Z0L21,MaMLFL22}]{
\includegraphics[width=.98\linewidth]{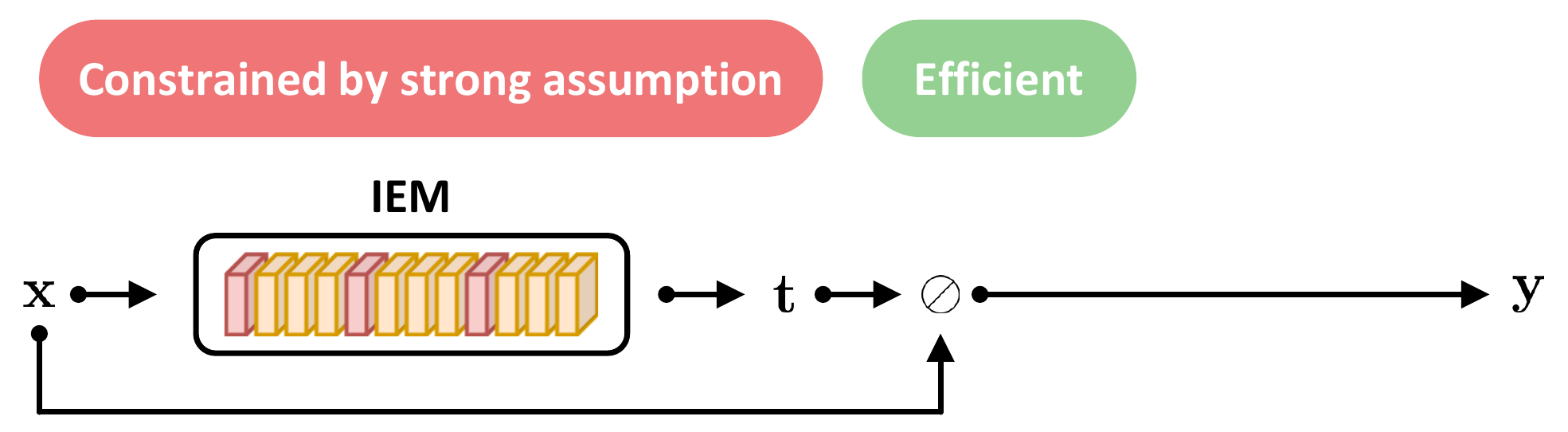}
\label{subf:illumination}}
\\
\subfloat[Our method, COmparametric Neural Enhancer (CONE)]{
\includegraphics[width=.98\linewidth]{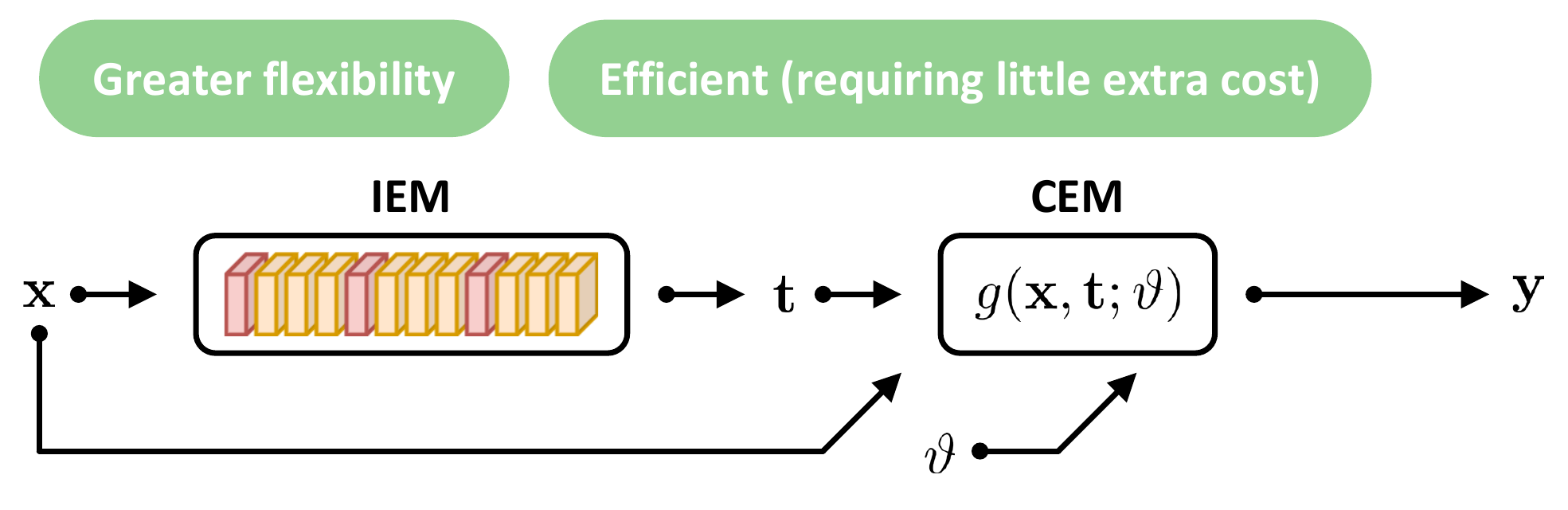}
\label{subf:cone}}
\caption{Comparison between conventional deep image enhancement and CONE. $\x$: low-light image. $\mathbf{c}$: image component. $\y$: enhanced image. $\t$: illumination map. $\vartheta$: CEM parameters. IEM: illumination estimation module. CEM: comparametric equation module.}
\label{f:novelty}
\end{figure}

Decompose-and-enhance methods~\cite{WeiWY018,XuYYL20,Yang0FW020,ZhangGMLZ21} break down a low-light image into two or more image components (e.g., illumination and reflectance, low and high frequency layers, multiscale band representations, etc.), improve image quality for each, and finally recompose them to recover an enhanced image (Fig.~\ref{subf:decompose-and-enhance}). Since deep neural networks (DNNs) are used for both decomposition and enhancement, these methods allow for high degree-of-freedom (DOF) image enhancement, but are usually less efficient for the same reason. In the meantime, illumination estimation-centric methods~\cite{WangZFSZJ19,ZhangDZW20,Liu0Z0L21,MaMLFL22} directly regard image reflectance (pixel-wise division of the low-light image by an illumination map) as the desired enhancement result, so require only one single DNN for image-to-illumination conversion (Fig.~\ref{subf:illumination}). These methods are more efficient, but the implicit assumption regarding the reflectance may be strong; it constrains image enhancement to a non-parametric transformation of the low-light image and illumination map, which cannot easily and flexibly adapt to the diverse demands of different applications. In this research, we aim to alleviate this constraint and discover a greater balance between the enhancement flexibility and the computation efficiency.

In 2000, Mann~\cite{Mann00} proposed comparametric equations that describe a parametric relationship between a camera response function (CRF) and a dilated version of the same function. Parameterized by the CRF and an exposure compensation ratio (called `exposure ratio' hereafter), comparametric equations can be used for conversion between differently exposed photos of the same scene. Recently, Ying et al.~\cite{YingLRWW17} proposed to define the exposure ratio as the reciprocal of image illumination and showed high efficacy in low-light image enhancement. These works enable flexible conversion between photos, but are handcrafted and not easily adaptable to different tasks.

In this research, we propose a trainable comparametric equation module (CEM) for diversified conversion from the low-light image and illumination map to the enhanced image (Fig.~\ref{subf:cone}). We embed this module in an illumination estimation-centric neural network and incorporate it in unsupervised learning, leading to a novel image enhancement method referred to as COmparametric Neural Enhancer (CONE). Unlike the handcrafted previous studies~\cite{Mann00,YingLRWW17,RenYLL19}, CONE trains CEM with task-dependent image enhancement losses and allows for joint learning with illumination estimation. It offers greater enhancement flexibility compared to illumination estimation-centric methods~\cite{WangZFSZJ19,ZhangDZW20,Liu0Z0L21,MaMLFL22} while inheriting their high efficiency\footnote{In this study, the term `flexibility' indicates the ability of a method to be easily adapted to different tasks (datasets, losses, etc).}. Experiments demonstrate the superiority of CONE over the state of the art.

\section{Proposed Method}
\label{s:proposed}

As shown in Fig.~\ref{subf:cone}, CONE consists of one illumination estimation module (IEM) and one comparametric equation module (CEM).

\subsection{Illumination Estimation Module}
\label{subs:iem}

Our CEM can be applied to any illumination estimation-centric neural networks. We adopt SCI~\cite{MaMLFL22} as the backbone network of the IEM because of its great inference efficiency.

Thus, the IEM consists of an enhancement network and a self-calibrated network. The enhancement network is the main part of the IEM, learning the mapping from a low-light image to an illumination map. It has four convolution blocks ($[3\times3,\ 3]$ Conv + ReLU) and one skip connection. The self-calibrated network is utilized only to aid in the training of the enhancement network and not involved in inference. It contains eight convolution blocks ($[3\times3,\ 16]$ Conv + BatchNorm + ReLU) and three skip connections. The output of the IEM is a spatially smooth illumination map of exactly the same size as the low-light image.

\subsection{Comparametric Equation Module}
\label{subs:cem}

According to the Retinex theory, a low-light image $\x$ can be formulated as $\x=\t\otimes\r$, where $\t$ and $\r$ are illumination and reflectance components and $\otimes$ denotes pixel-wise multiplication. Illumination estimation-centric methods~\cite{WangZFSZJ19,ZhangDZW20,Liu0Z0L21,MaMLFL22} directly regard $\r$ as the desired enhancement result $\y$:
\begin{equation}
\y=\r=\x\oslash\t
\label{e:reflectance}
\end{equation}
with $\oslash$ being pixel-wise division. If $\x$ and $\t$ are fixed, this pixel-wise division is less flexible because it is non-parametric and hard to adapt to different image enhancement tasks.
In this study, we incorporate the intermediate CEM in our network to associate the illumination map $\t$ with the desired enhancement result $\y$ in a parameterized way.

\begin{figure}[t]
\centering
\includegraphics[width=.98\linewidth]{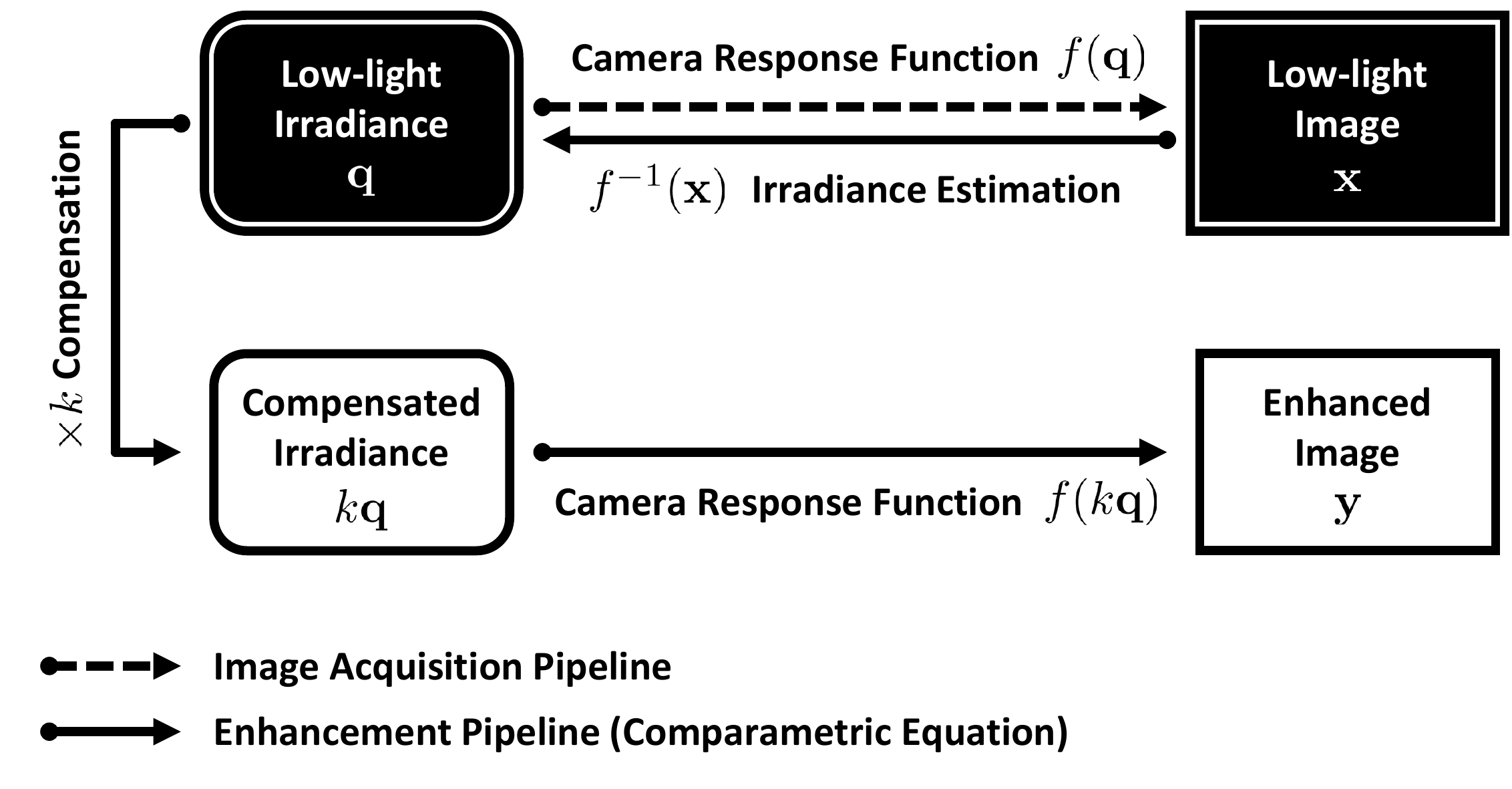}
\caption{Quantigraphic image enhancement via comparametric equation.}
\label{f:comparametric}
\end{figure}

\smallskip\noindent
\textbf{Comparametric Equation.} Fig.~\ref{f:comparametric} shows the idea of comparametric equations~\cite{Mann00}. As shown by the dashed arrow, digital image acquisition via a camera can be understood very roughly as a mapping by CRF $f(\q)$ from scene irradiance $\q$ to digital image $\x$. Suppose that this CRF is known. Given a low-light image $\x$, the irradiance $\q$ can be inversely estimated through $f^{-1}(\x)$. Compared to the enhancement of the low-light image, which is in general highly non-linear, it is easier to compensate the irradiance because the latter varies linearly with the exposure time. Let $k$ be a desired exposure ratio (scale factor of exposure time). The irradiance $\q$ can thus be compensated by $k\q$. In consequence, an enhanced image $\y$ can be generated by passing the compensated irradiance $k\q$ through the CRF $f(k\q)$. The above processes from the low-light image $\x$ to the enhanced image $\y$ form an enhancement pipeline (solid arrows in Fig.~\ref{f:comparametric}), and can be mathematically expressed by a comparametric equation:
\begin{equation}
\y=f(kf^{-1}(\x)).
\label{e:comparametric-1}
\end{equation}
Parameterized by $f(\cdot)$ and $k$, Eq.~\eqref{e:comparametric-1} can approximate more flexible conversions (than Eq.~\eqref{e:reflectance}) between differently exposed photos. We thus employ it as the core of the CEM. At stake is how to determine the CRF and the exposure ratio.

\smallskip\noindent
\textbf{Exposure Ratio.} Borrowing the idea from Ying et al.~\cite{YingLRWW17}, instead of defining $k$ as a scalar identical for all pixels, we formulate it as a matrix representing the desired exposure ratio that varies spatially for each pixel. Specifically, this matrix $\k$ is defined as a positive real matrix inversely proportional to the illumination map:
\begin{equation}
\k=1\oslash\t.
\label{e:exposure-ratio}
\end{equation}
Eq.~\eqref{e:exposure-ratio} is physically meaningful: to deliver the desired exposure, the darker areas of irradiance should be compensated to a larger extent than the brighter areas.

\begin{table}[t]
\centering
\caption{Comparametric equations exploited in this study.}
\label{t:comparametric}
\small
\begin{tabular*}{\linewidth}{@{\extracolsep{\fill}}ll}
\toprule
\textbf{Name} & \textbf{Equation} \\
\midrule
BetaGamma Correction (BGC) & $\y=e^{b(1-\k^{a})}\x^{\k^{a}}$\\
\midrule
Preferred Correction (PC) & $\y=\frac{\k^{ab}\x}{((\k^{a}-1)\x^{1/b}+1)^{b}}$ \\
\midrule
Sigmoid Correction (SC) & $\y=\frac{(b+1)\k^{a}\x}{(\k^{a}-1)\x+b+1}$ \\
\bottomrule
\end{tabular*}
\end{table}

\smallskip\noindent
\textbf{Camera Response Function.} There have been a number of existing functional forms proposed for CRF approximation. With Eq.~\eqref{e:comparametric-1} in mind, we prefer the CRF to be easily invertible. In order for training the CEM jointly with the IEM using stochastic gradient descent, it is also desirable for the CRF to be readily differentiable. In CONE, we take account of three CRF functional forms proposed by Mann~\cite{Mann00} and Eilertsen et al.~\cite{EilertsenKDMU17}. The comparametric equations derived from them~\cite{Mann00,YingLRWW17,RenYLL19} are shown in Table~\ref{t:comparametric}. All these equations contain two parameters $a$ and $b$.

Let $\vartheta=(a,b)$ denote the set of these parameters. As illustrated in Fig.~\ref{subf:cone}, the comparametric equation in Eq.~\eqref{e:comparametric-1} is actually a function of two variables $\x$ and $\t$ with a set of parameters $\vartheta$, and can be rewritten as Eq.~\ref{e:comparametric-2}. It is more flexible than Eq.~\ref{e:reflectance} because the conversion from $\x$ and $\t$ to $\y$ can be learned from data by optimizing the parameter $\vartheta$.
\begin{equation}
\y=g(\x,\t;\vartheta).
\label{e:comparametric-2}
\end{equation}
Most previous studies on CRF approximation learns $\vartheta$ through curve fitting, interpolation, or regression~\cite{Mann00,GrossbergN04,EilertsenKDMU17,YingLRWW17,RenYLL19} on real-world camera response curves. However, the collection of these curves is far more difficult than for image due to trade secret issues, making it difficult to adapt to different image enhancement tasks. In this study, we directly train CEM using unsupervised image enhancement losses for greater task adaptability.

\subsection{Unsupervised Learning}
\label{subs:loss}

We first follow the practice of SCI~\cite{MaMLFL22} and define a loss on the illumination map as $\L^{(\t)}=\L_{sm}+\omega_{f}\L_{f}$, where $\L_{sm}$ and $\L_{f}$ denote a smoothness loss and a fidelity loss, respectively, with $\omega_{f}=1.5$ controlling the balance between $\L_{sm}$ and $\L_{f}$.
The smoothness loss is used to force $\t$ to possess an edge-preserving smoothness property. The fidelity loss is to guarantee the pixel-wise consistency between the input $\x$ and $\t$. More details can be found in the original paper.

Note that $\L^{(\t)}$ is totally independent on the enhanced image $\y$, so has no effect on the training of CEM. To enable the learning of $\vartheta$, we define an additional unsupervised learning loss on $\y$, inspired by ZeroDCE++~\cite{LiGL22}:
\begin{equation}
\L^{(\y)}=\L_{e}+\L_{sp}+\omega_{c}\L_{c}.
\label{e:loss-y}
\end{equation}
Here, $\L_{e}$, $\L_{sp}$, and $\L_{c}$ are exposure control loss, spatial consistency loss, and color constancy loss, respectively. We empirically set $\omega_{c}=0.5$ in all experiments, which controls the contribution of $\L_{c}$.

\begin{table}[t]
\centering
\caption{Performance of models w/o and w/ CEM. `IEM+CEM' w/o `$\star$' fixes $\vartheta$ and sets them to the same values used in LECARM~\cite{RenYLL19}; `$\star$ IEM+CEM' jointly learns $\vartheta$ with IEM. Best performance in bold.}
\label{t:cem}
\small
\begin{tabular*}{\linewidth}{@{\extracolsep{\fill}}lcccc}
\toprule
& \multicolumn{2}{c}{\textbf{MIT~\cite{BychkovskyPCD11}}} & \multicolumn{2}{c}{\textbf{LSRW~\cite{HaiXYHZLH21}}} \\
\cmidrule{2-3}
\cmidrule{4-5}
\textbf{Method} & \textbf{PSNR$\uparrow$} & \textbf{SSIM$\uparrow$} & \textbf{PSNR$\uparrow$} & \textbf{SSIM$\uparrow$} \\
\midrule
IEM & 19.87 & 0.836 & 14.08 & 0.388 \\
\midrule
IEM+CEM (BGC) & 16.31 & 0.759 & 11.82 & 0.319 \\
IEM+CEM (PC) & 14.66 & 0.733 & 13.20 & 0.378 \\
IEM+CEM (SC) & 16.15 & 0.774 & 14.54 & 0.449 \\
\midrule
$\star$ IEM+CEM (BGC) & 17.44 & 0.777 & \textbf{17.39} & 0.460 \\
$\star$ IEM+CEM (PC) & 18.47 & 0.775 & 17.20 & \textbf{0.468} \\
$\star$ IEM+CEM (SC) & \textbf{21.19} & \textbf{0.853} & 14.78 & 0.409 \\
\bottomrule
\end{tabular*}
\end{table}

\smallskip\noindent
\textbf{Exposure Control Loss.} This loss is defined to moderate under and over-exposure and measures the pixel-wise distance between $\y$ and a desired exposure level $\epsilon$ ($0.6$ as in ZeroDCE++):
\begin{equation}
\L_{e}=\frac{1}{n}\|\y-\epsilon\|_{2}^{2}
\label{e:loss-e}
\end{equation}
with $n$ being the number of pixels. Note that a 2D average pooling is applied over $\y$, with the kernel size being $16\times16$, before computing Eq.~\eqref{e:loss-e}.

\smallskip\noindent
\textbf{Spatial Consistency Loss.} This loss promotes spatial coherence of $\x$ and $\y$, capturing their image gradient difference:
\begin{equation}
\L_{sp}=\frac{1}{n}\|\nabla\y-\nabla\x\|_{2}^{2}
\label{e:loss-sp}
\end{equation}
with $\nabla$ denoting the first-order derivatives. Similar to Eq.~\eqref{e:loss-e}, a 2D average pooling with a kernel size of $4\times4$ is applied to both $\x$ and $\y$ before computing Eq.~\eqref{e:loss-sp}.

\smallskip\noindent
\textbf{Color Constancy Loss.} The gray-world color constancy hypothesis assumes that in a color-balanced photograph, the average of all colors is neutral gray. Based on this assumption, a color constancy loss is used to cancel out potential color deviations in $\y$:
\begin{equation}
\L_{c}=\sum\nolimits_{(i,j)}(\bar{y}_{i}-\bar{y}_{j})^{2}.
\label{e:loss-c}
\end{equation}
Here, $\bar{y}_i$ denotes the $i$-channel average intensity of the enhanced image and $(i,j)\in\{\mathrm{(R,G),(R,B),(G,B)}\}$.

\section{Experiments}
\label{s:experiments}

\begin{figure}[t]
\centering
\subfloat[Input]{
\includegraphics[height=.21\linewidth]{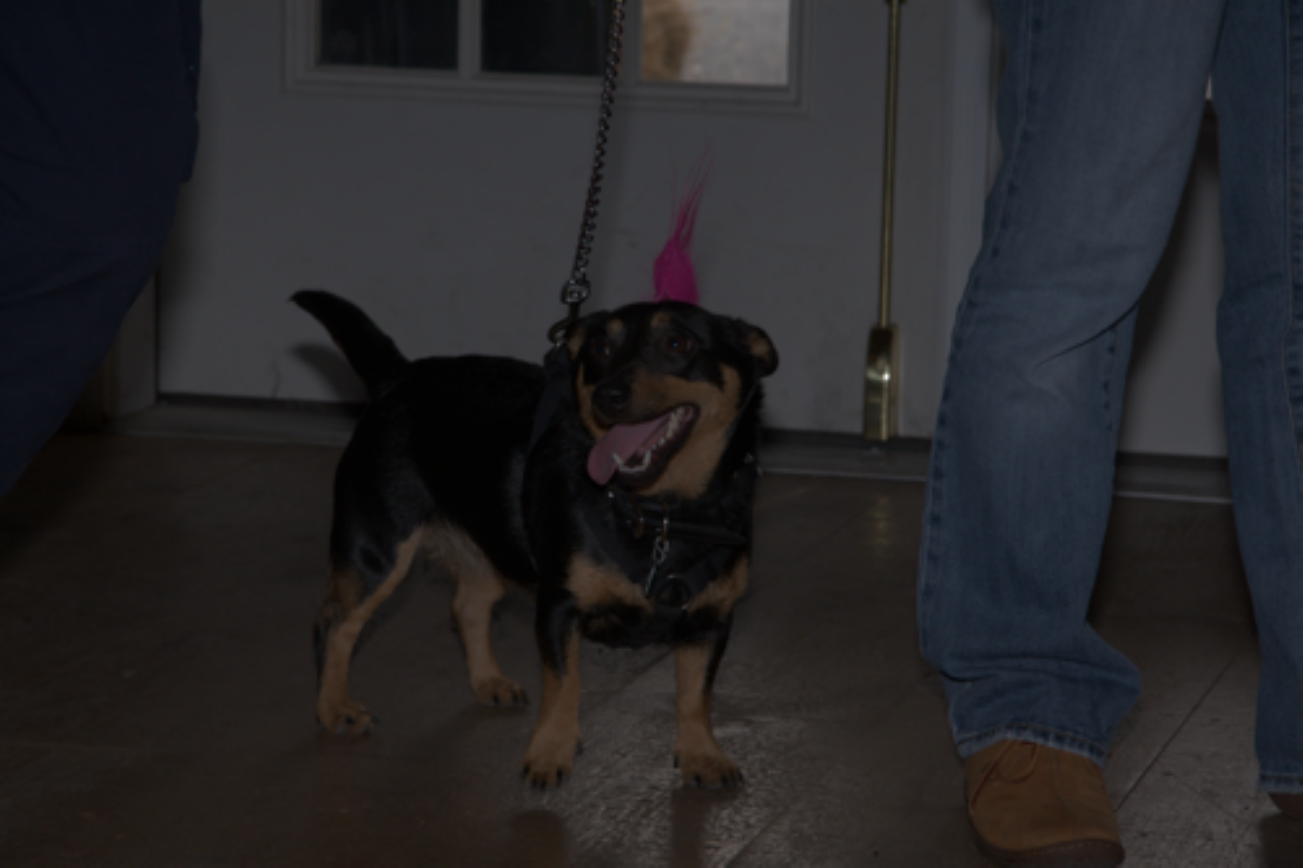}
\hfill
\includegraphics[height=.21\linewidth]{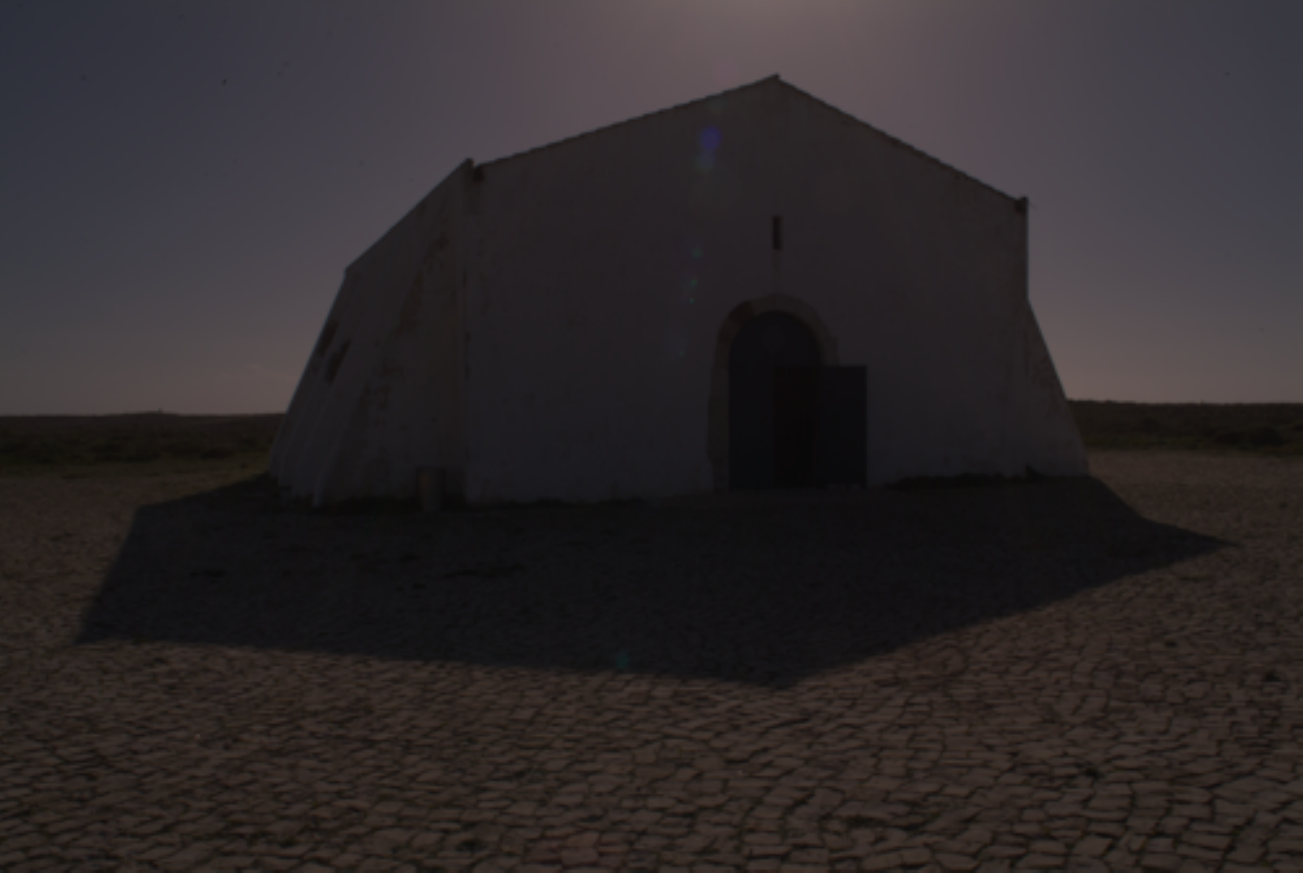}
\hfill
\includegraphics[angle=90,height=.21\linewidth]{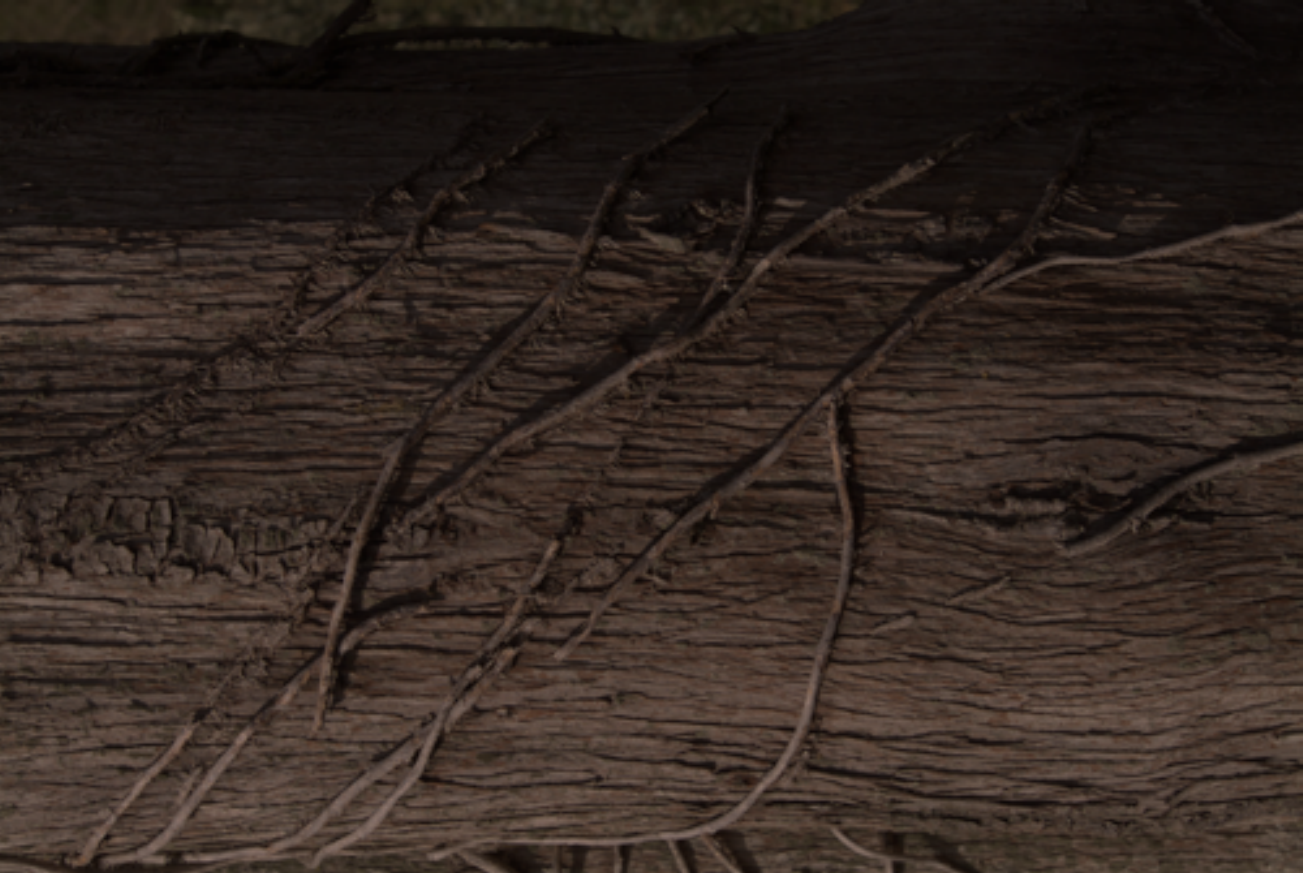}
\label{subf:input}}
\\
\subfloat[Enhanced image (IEM)]{
\includegraphics[height=.21\linewidth]{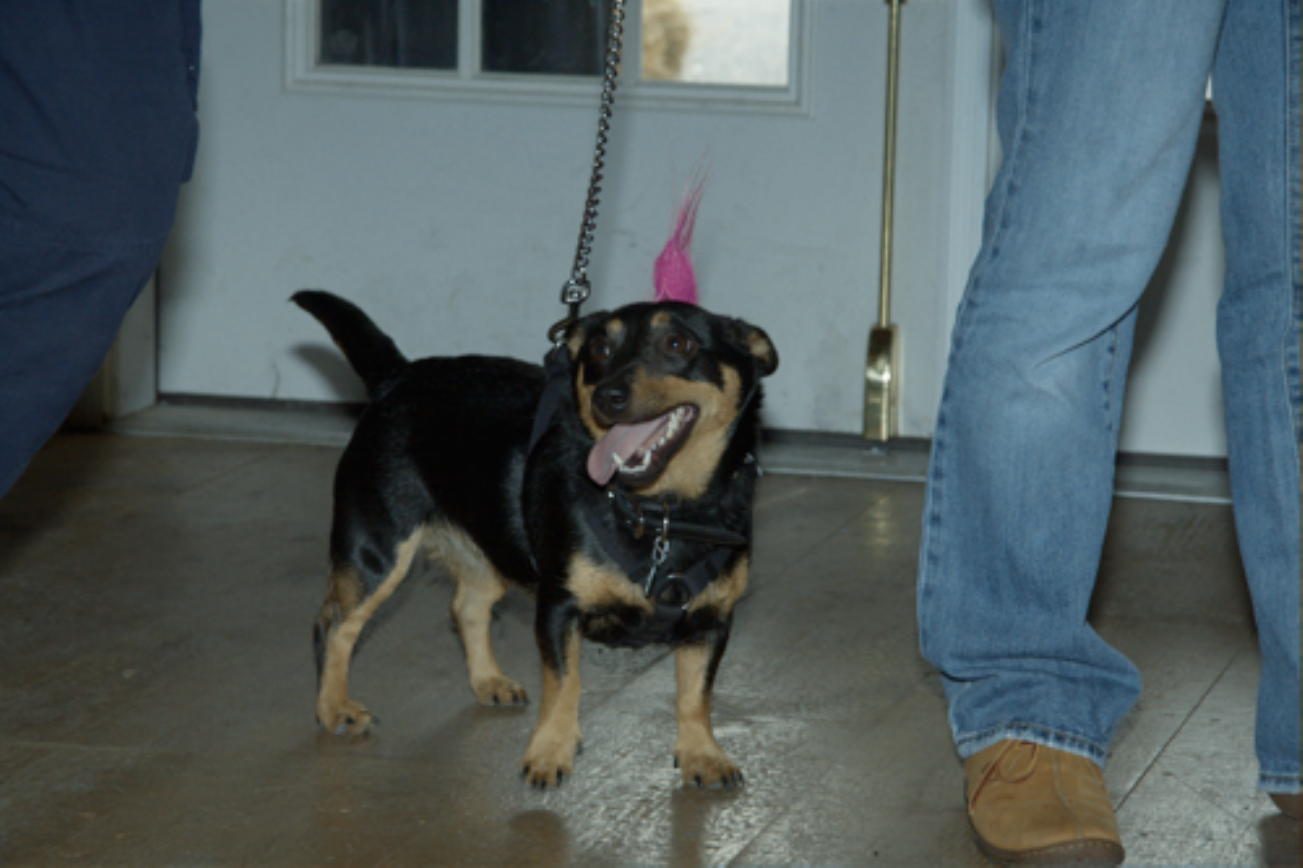}
\hfill
\includegraphics[height=.21\linewidth]{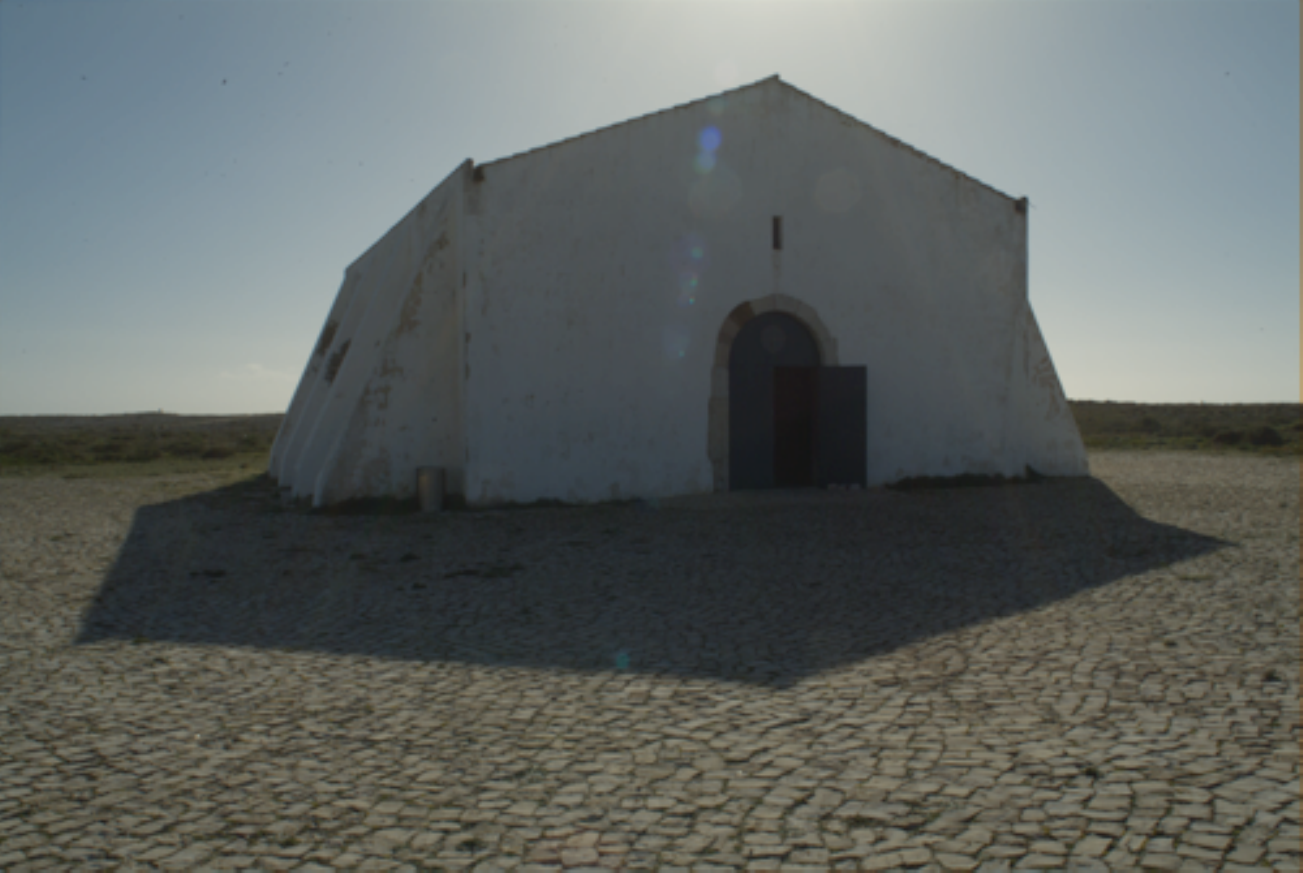}
\hfill
\includegraphics[angle=90,height=.21\linewidth]{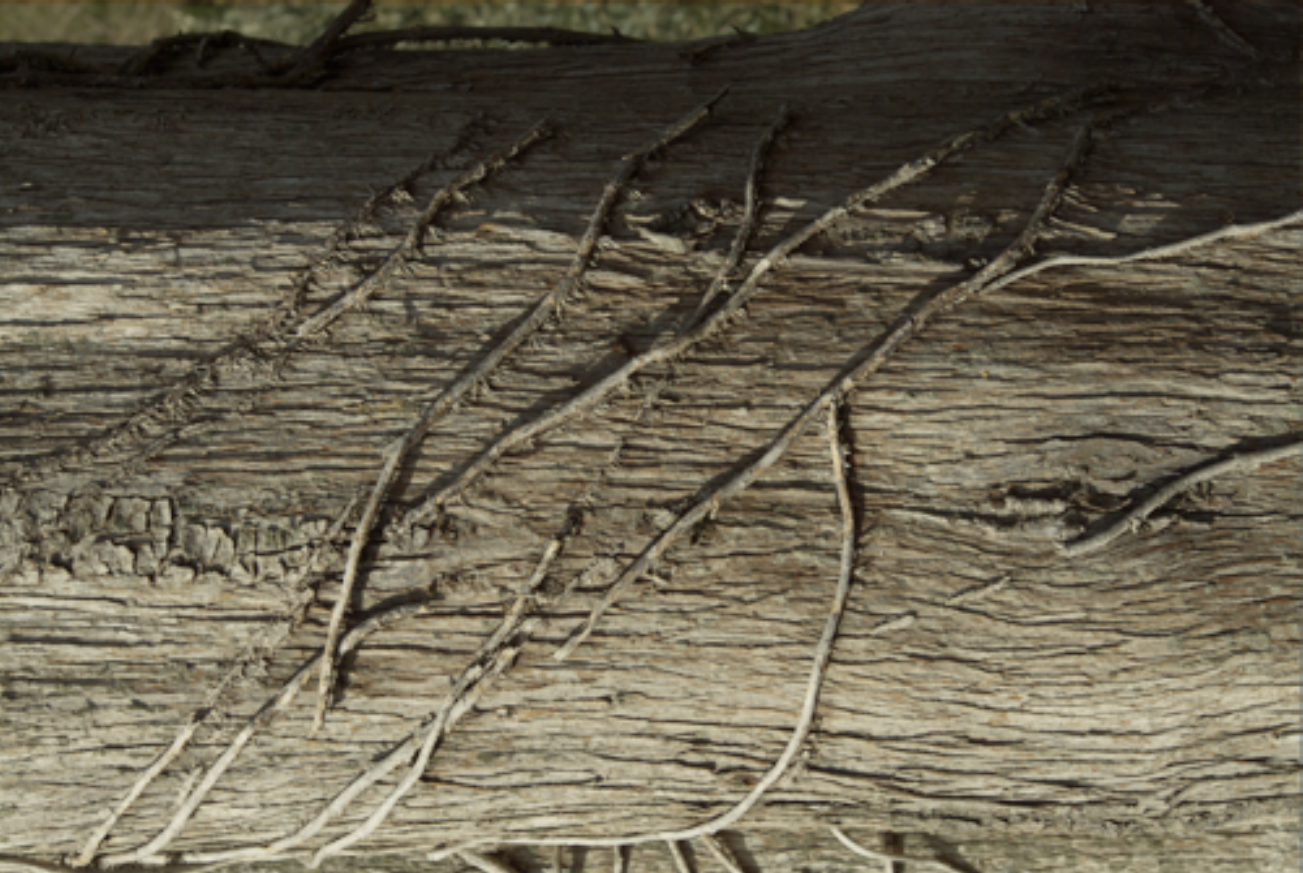}
\label{subf:baseline}}
\\
\subfloat[Error map (IEM)]{
\includegraphics[height=.21\linewidth]{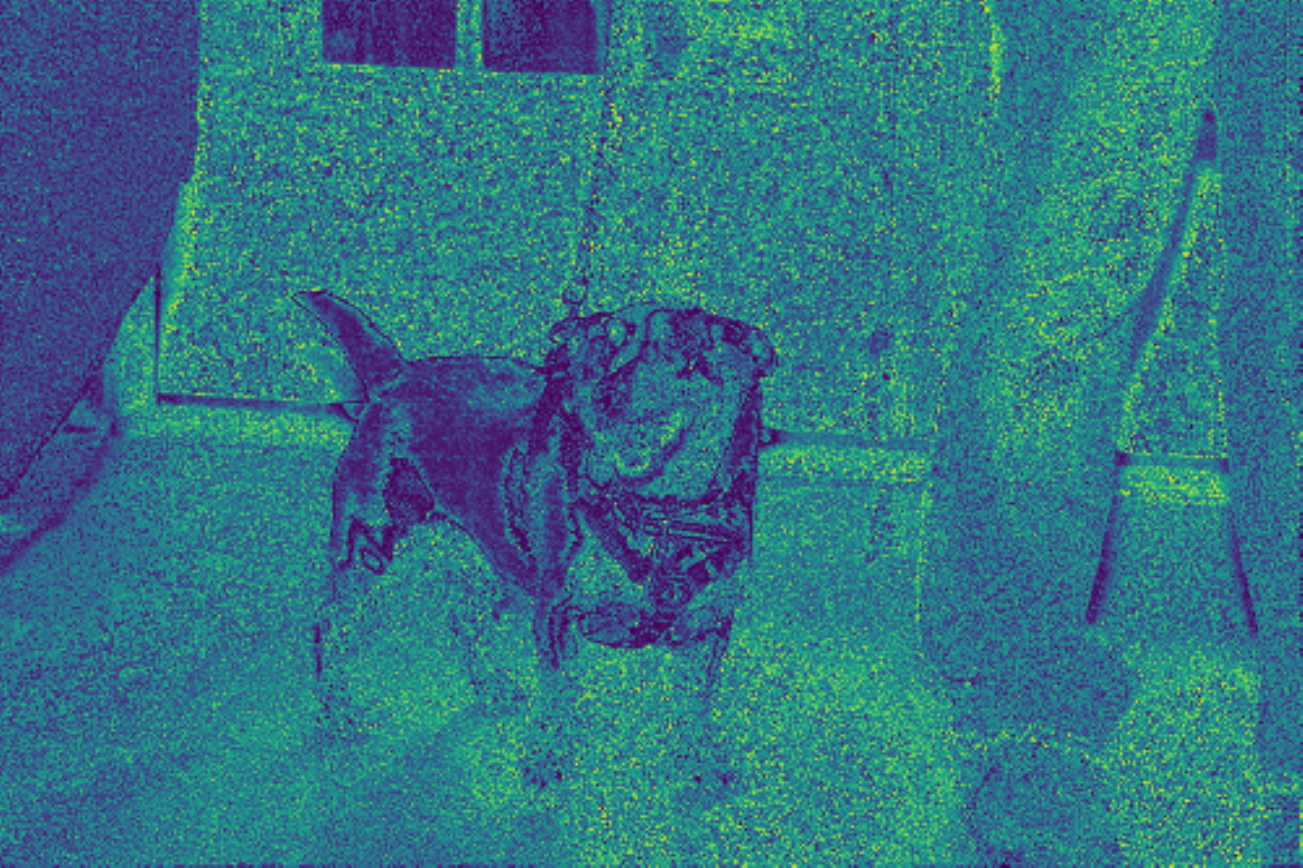}
\hfill
\includegraphics[height=.21\linewidth]{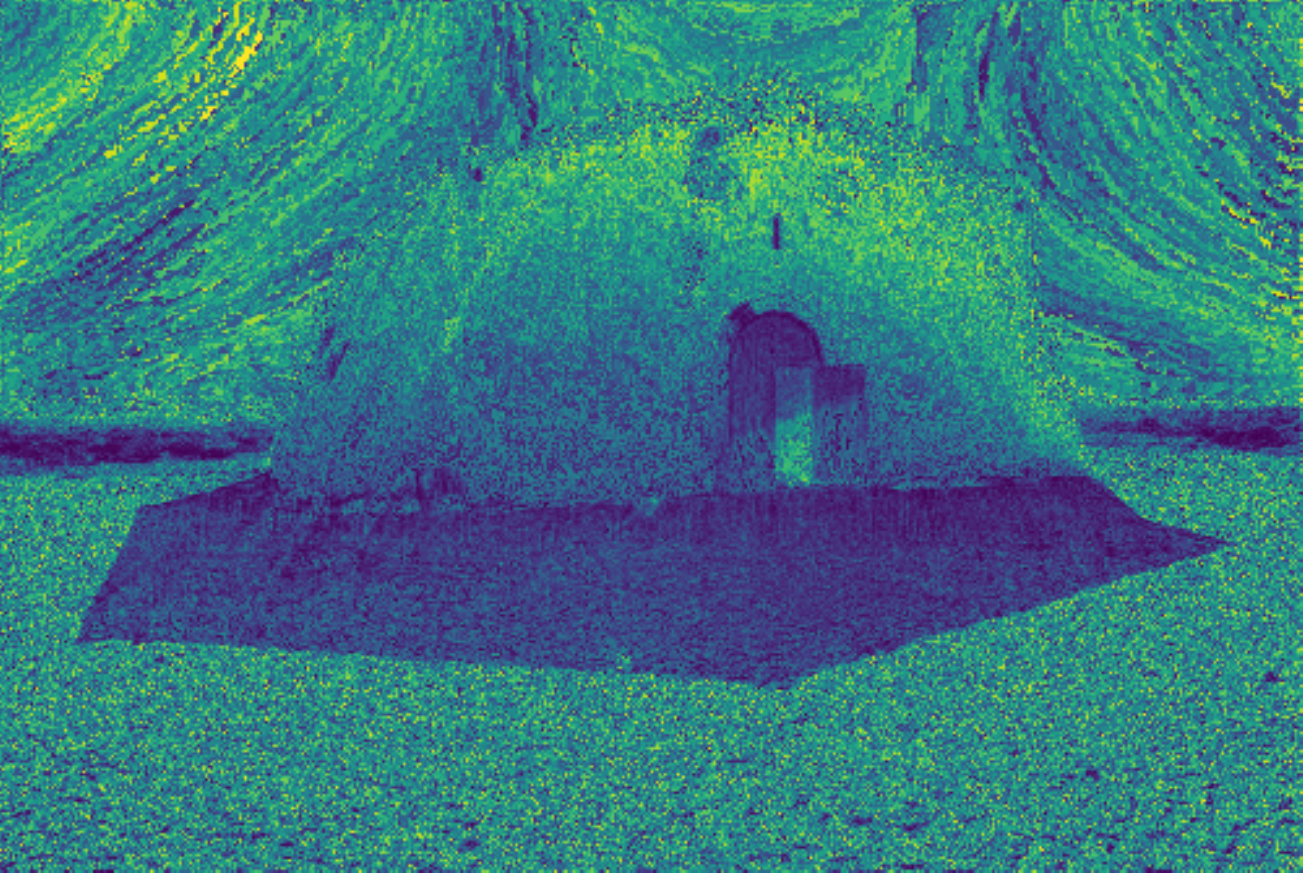}
\hfill
\includegraphics[angle=90,height=.21\linewidth]{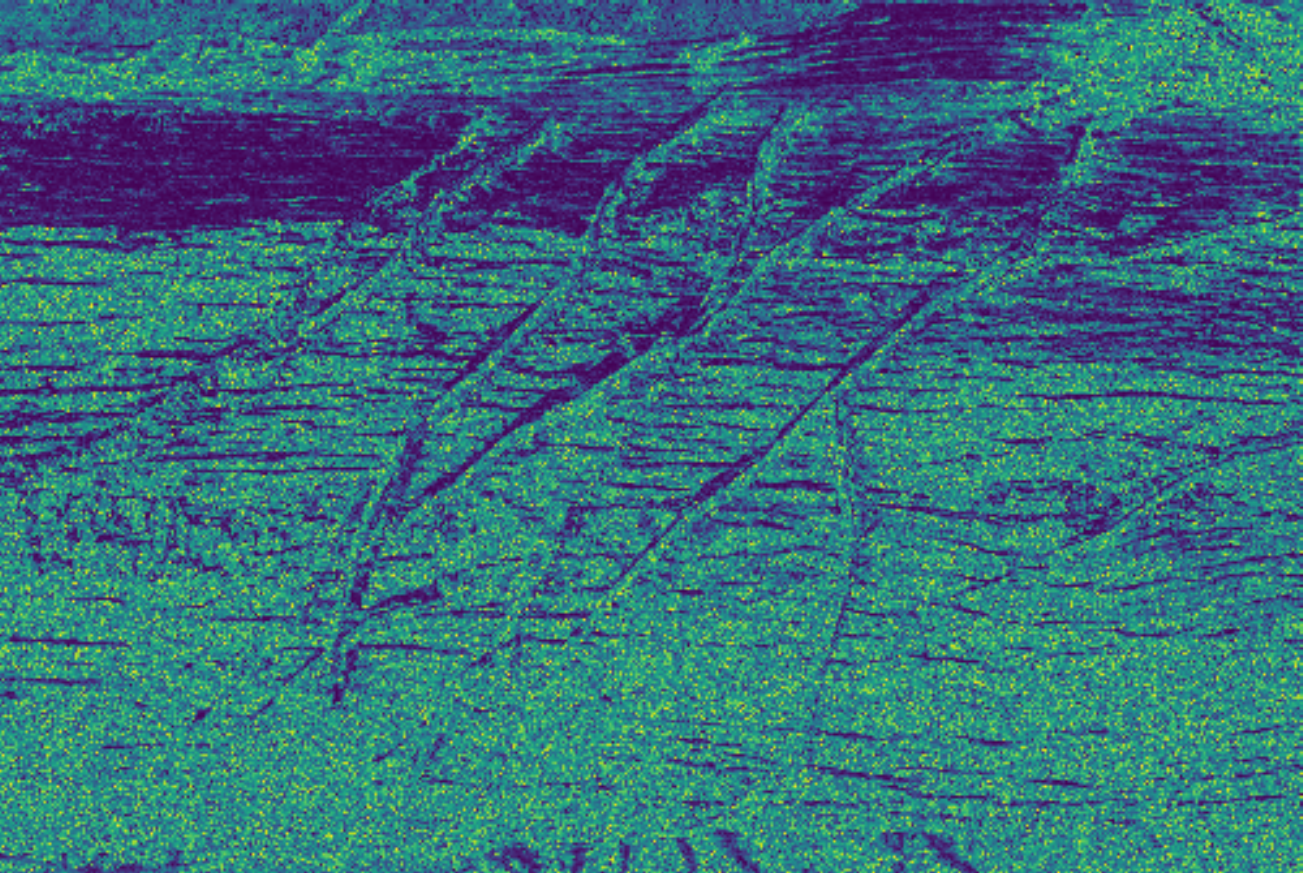}
\label{subf:baseline-diff}}
\\
\subfloat[Enhanced image {(IEM+CEM (SC))}]{
\includegraphics[height=.21\linewidth]{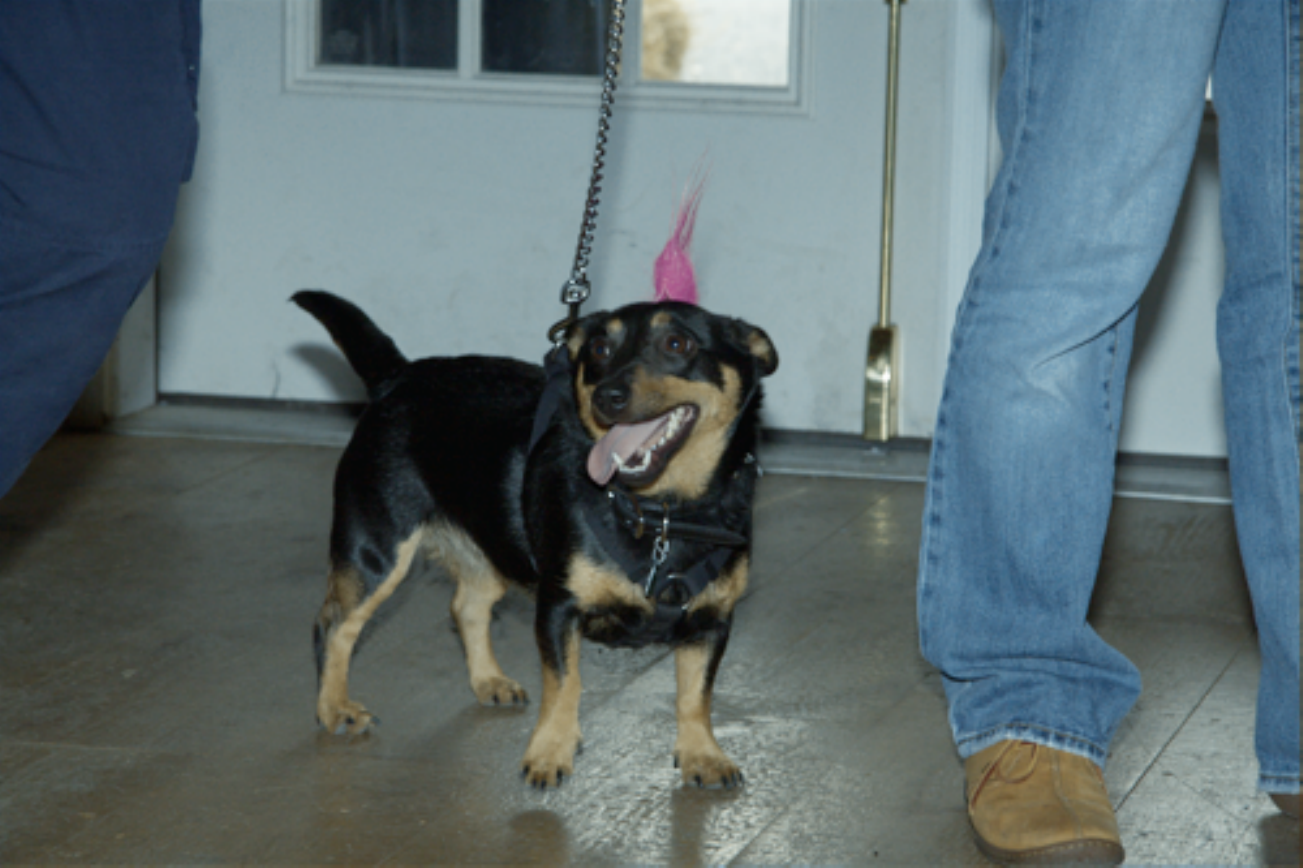}
\hfill
\includegraphics[height=.21\linewidth]{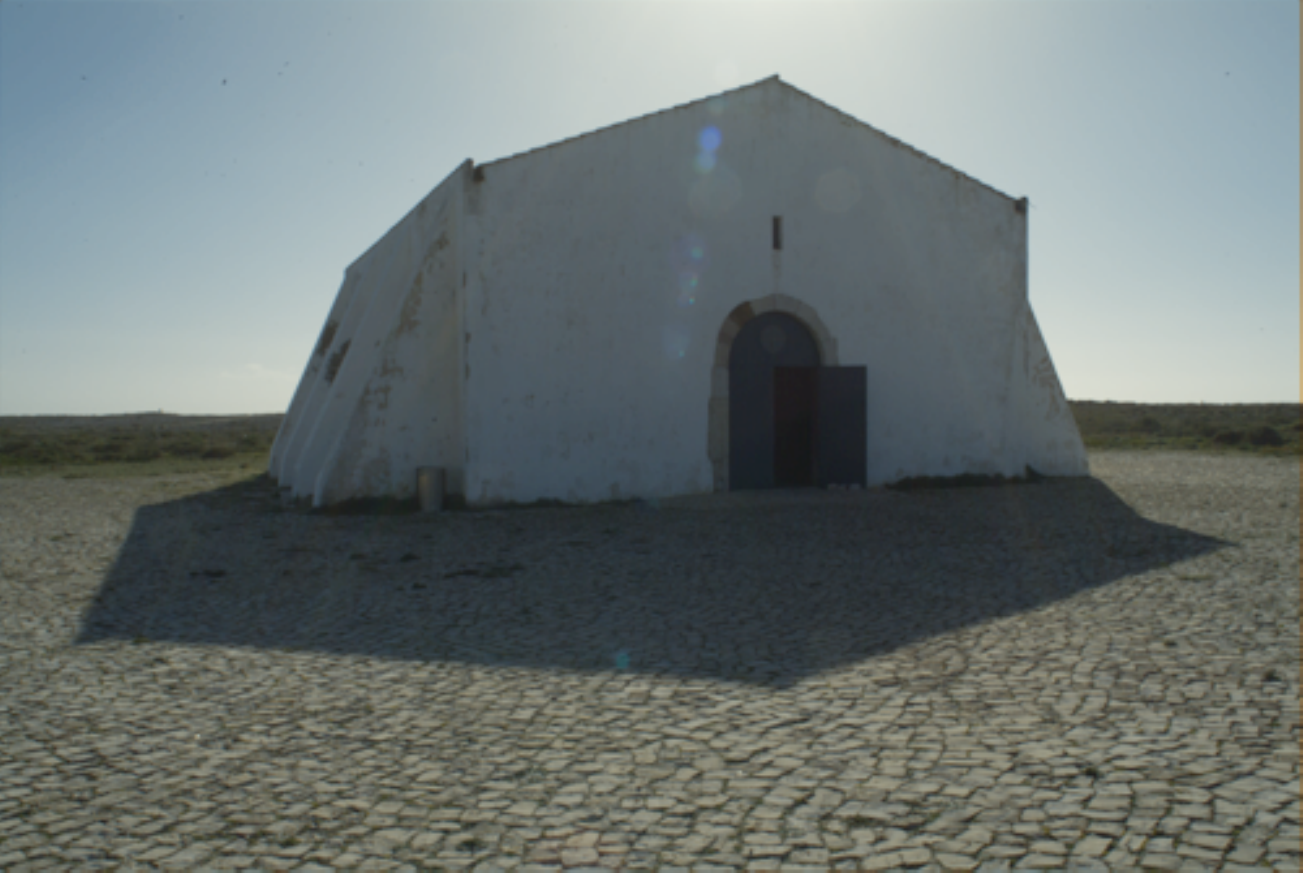}
\hfill
\includegraphics[angle=90,height=.21\linewidth]{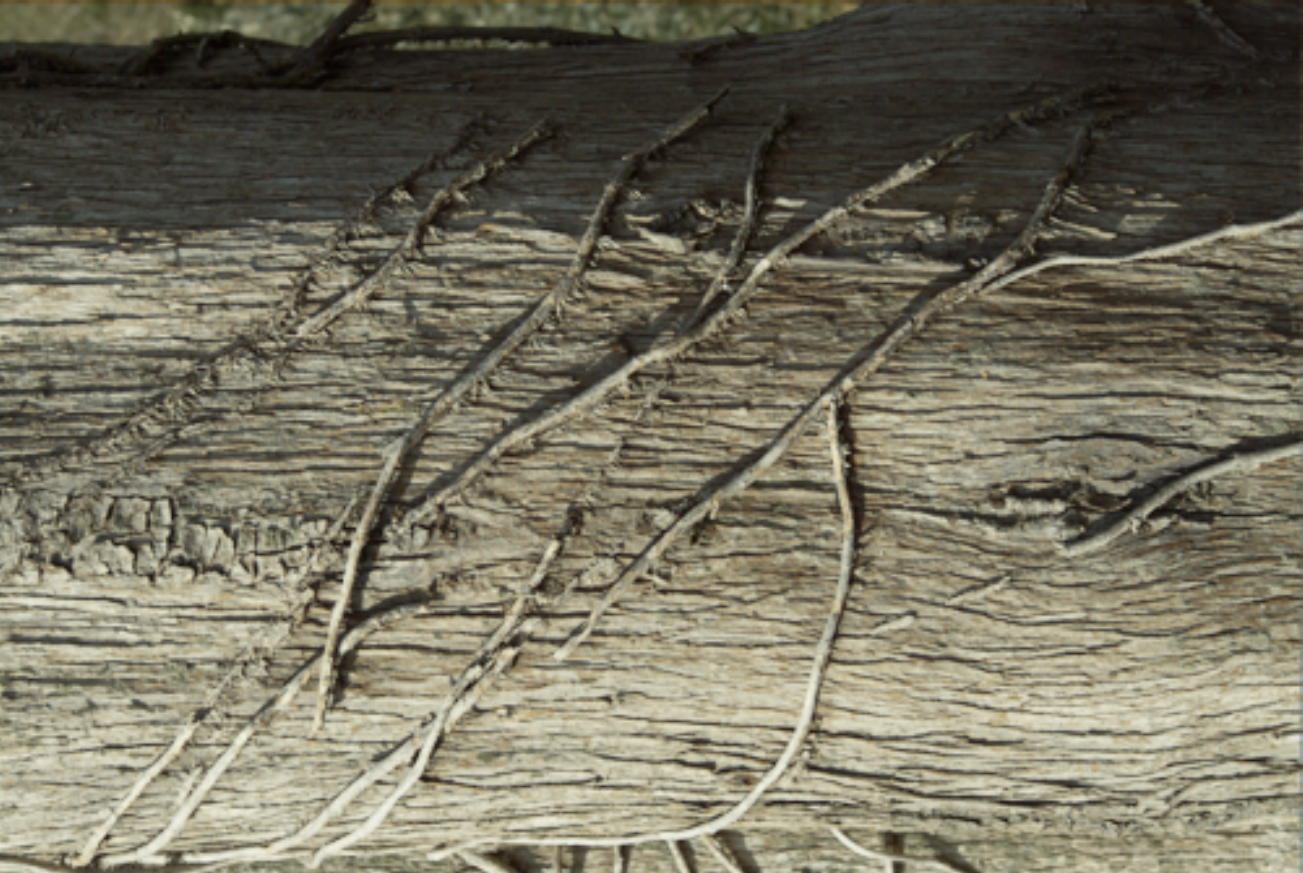}
\label{subf:visual-cone}}
\\
\subfloat[Error map {(IEM+CEM (SC))}]{
\includegraphics[height=.21\linewidth]{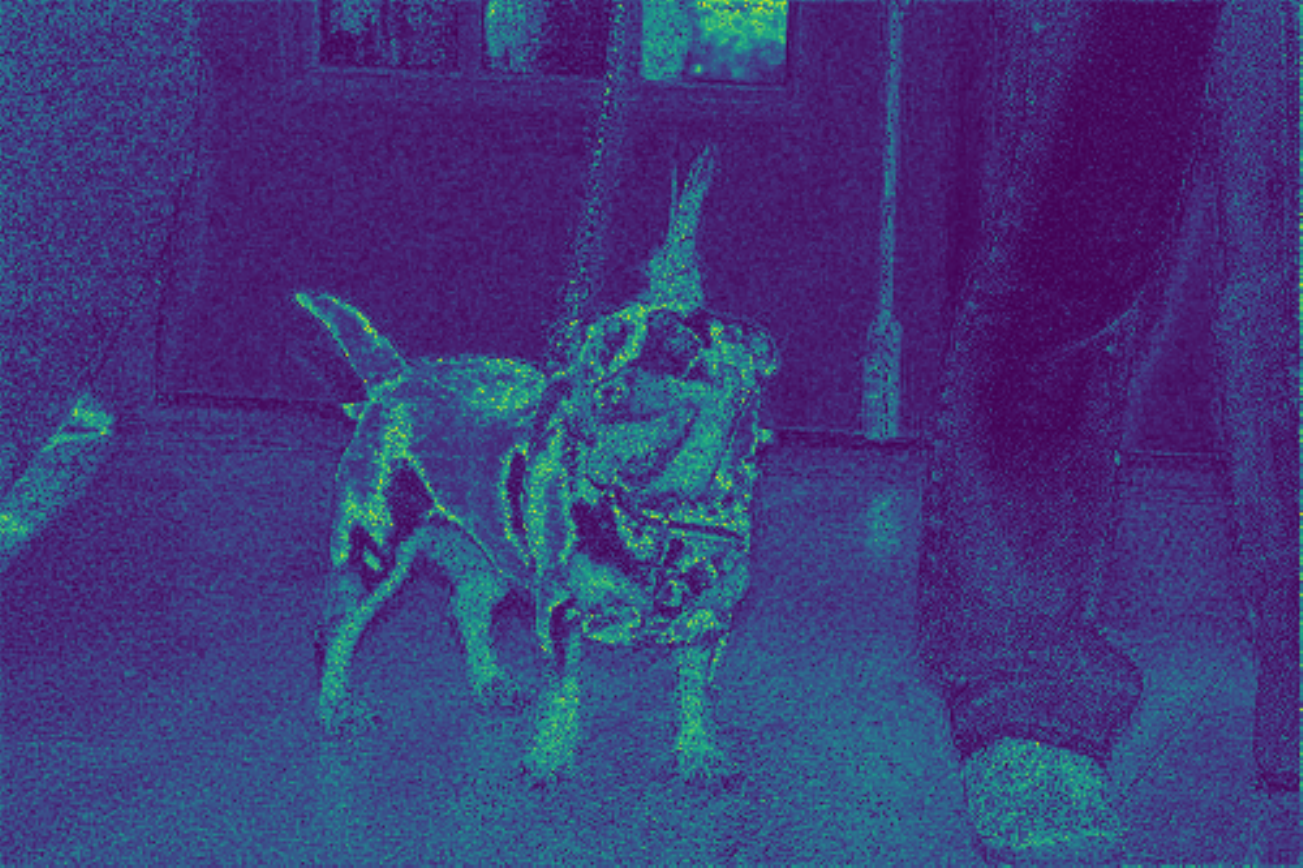}
\hfill
\includegraphics[height=.21\linewidth]{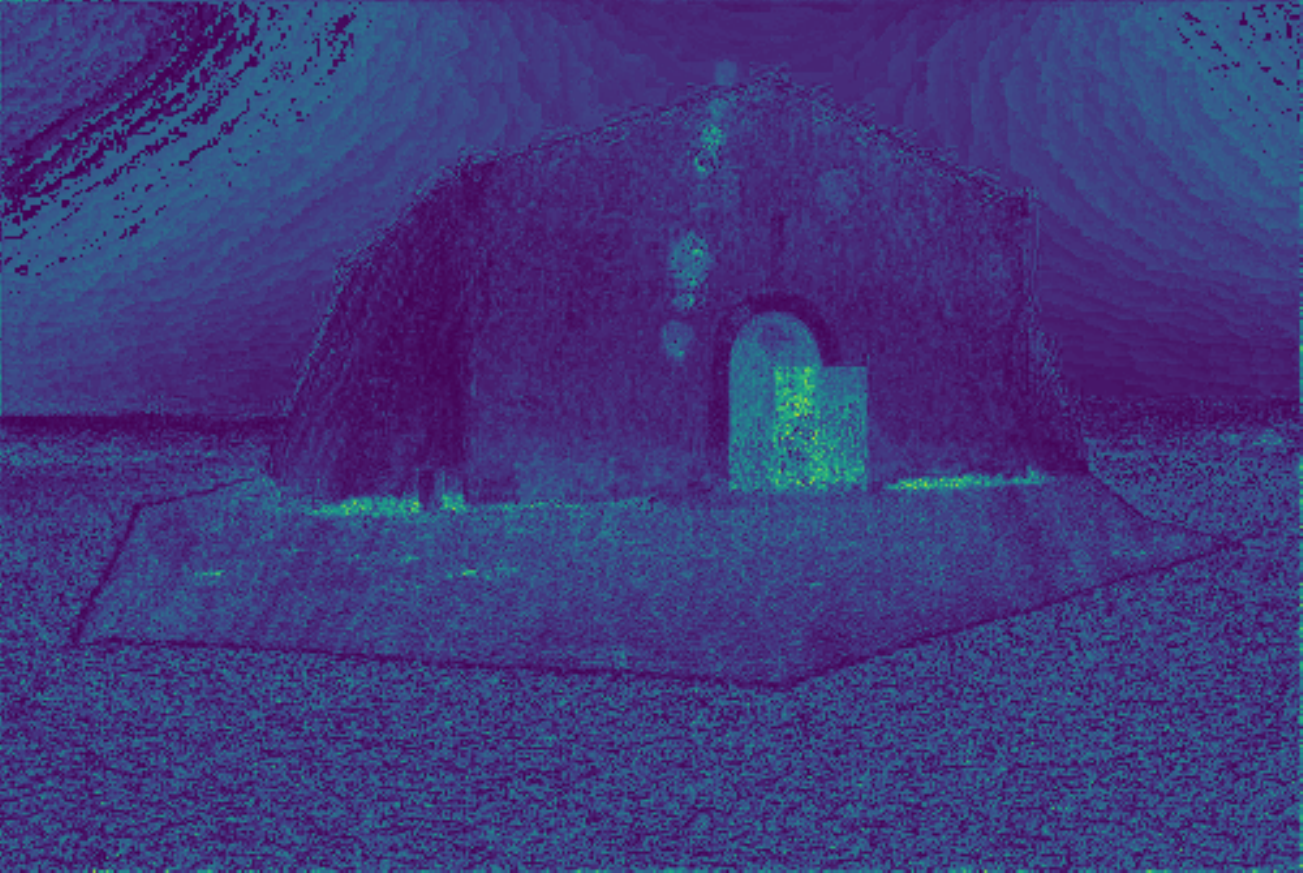}
\hfill
\includegraphics[angle=90,height=.21\linewidth]{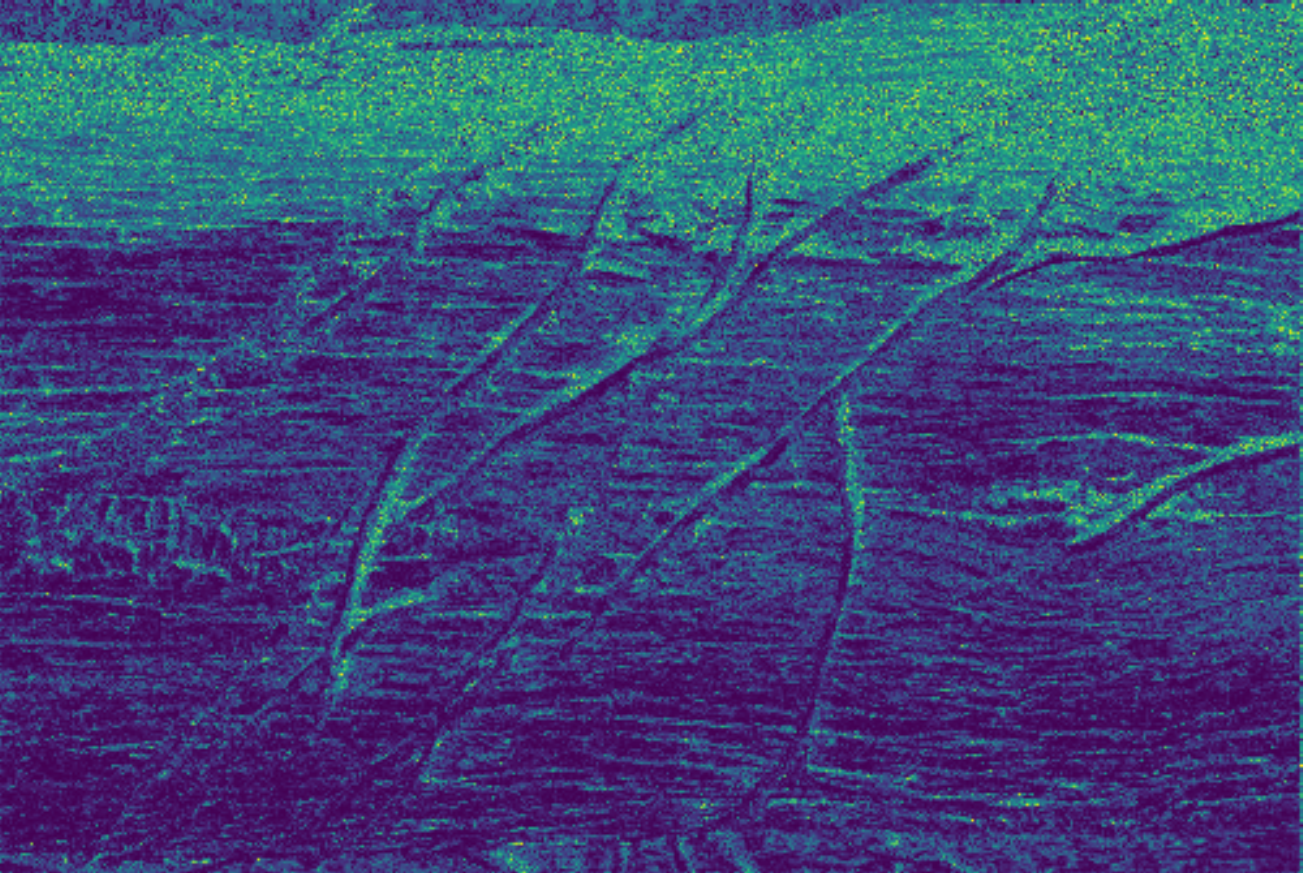}
\label{subf:visual-cone-diff}}
\\
\subfloat[Ground truth]{
\includegraphics[height=.21\linewidth]{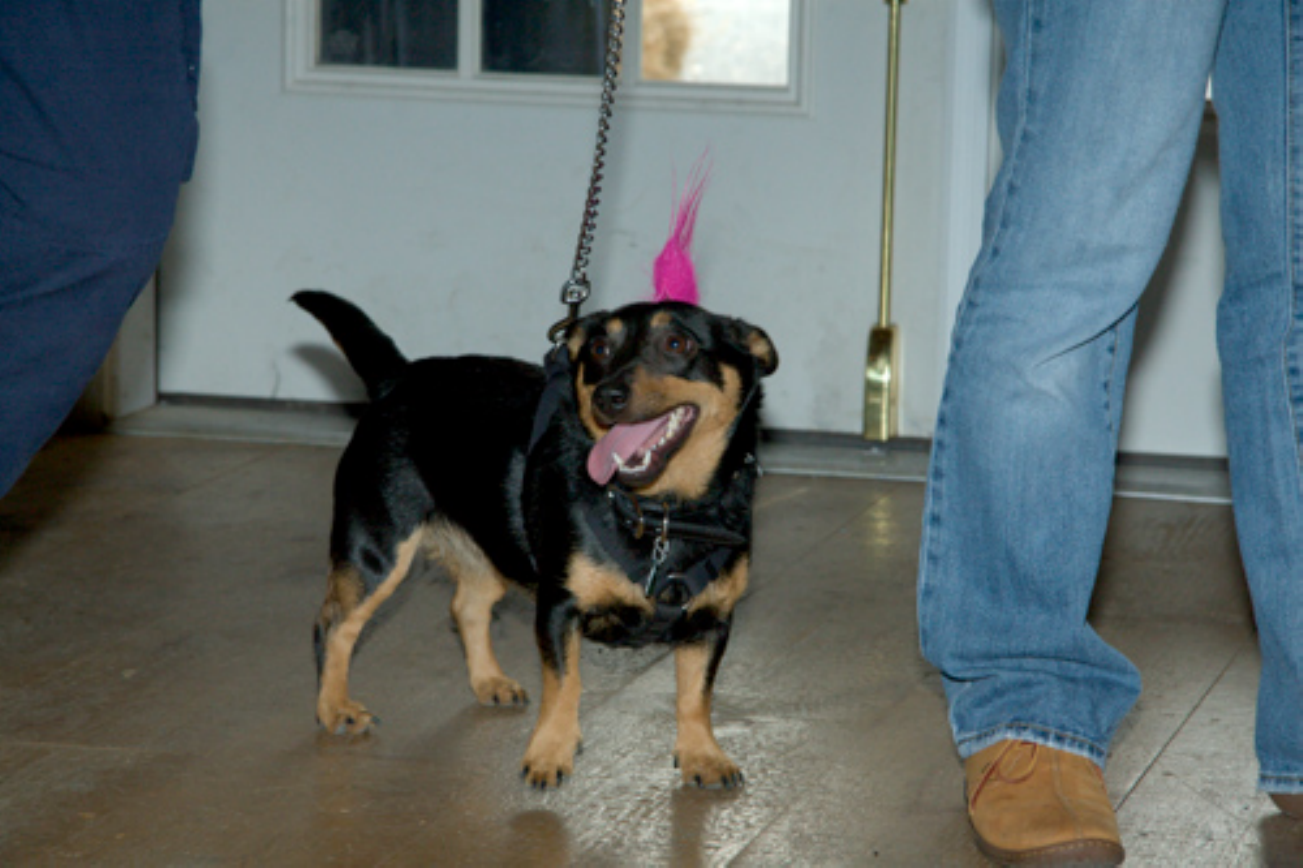}
\hfill
\includegraphics[height=.21\linewidth]{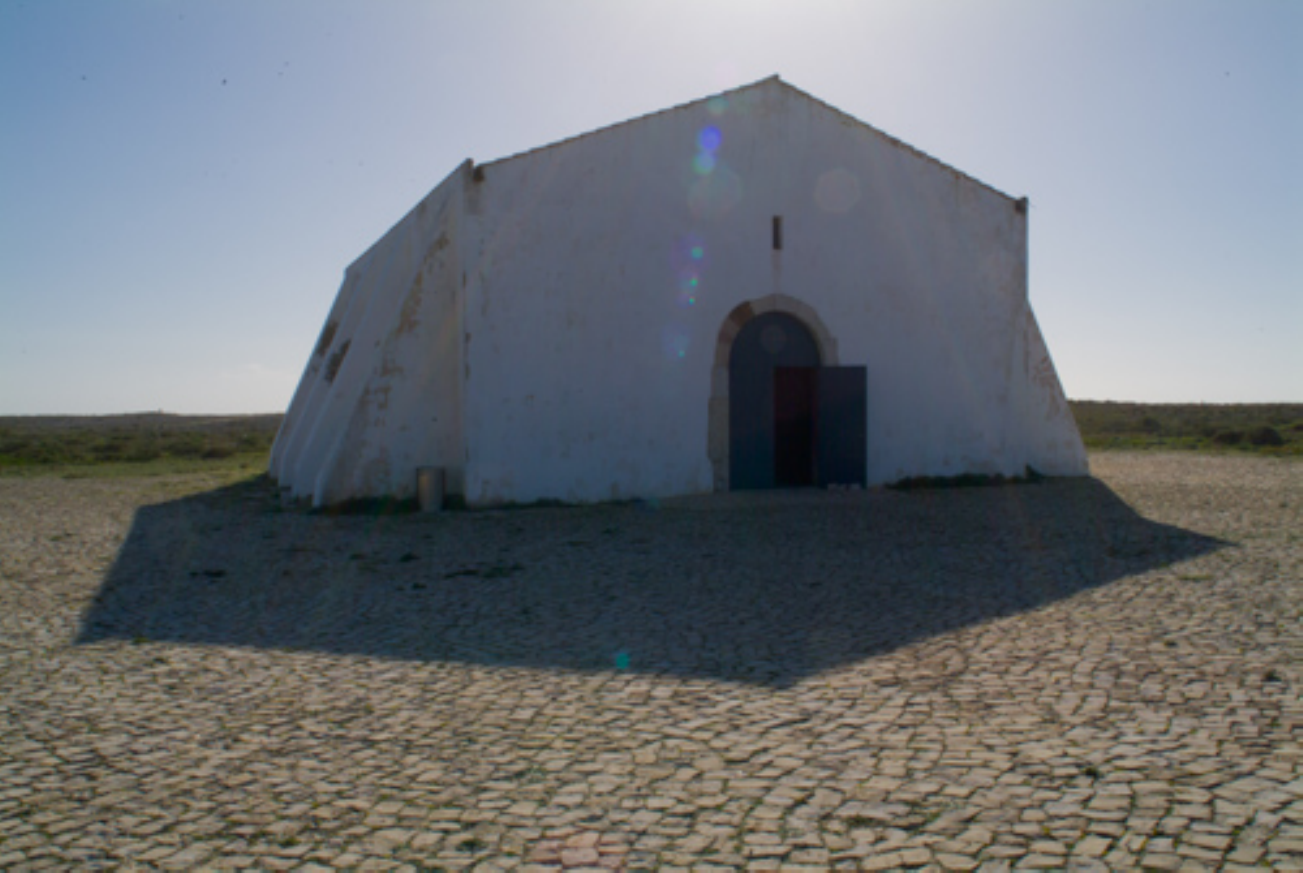}
\hfill
\includegraphics[angle=90,height=.21\linewidth]{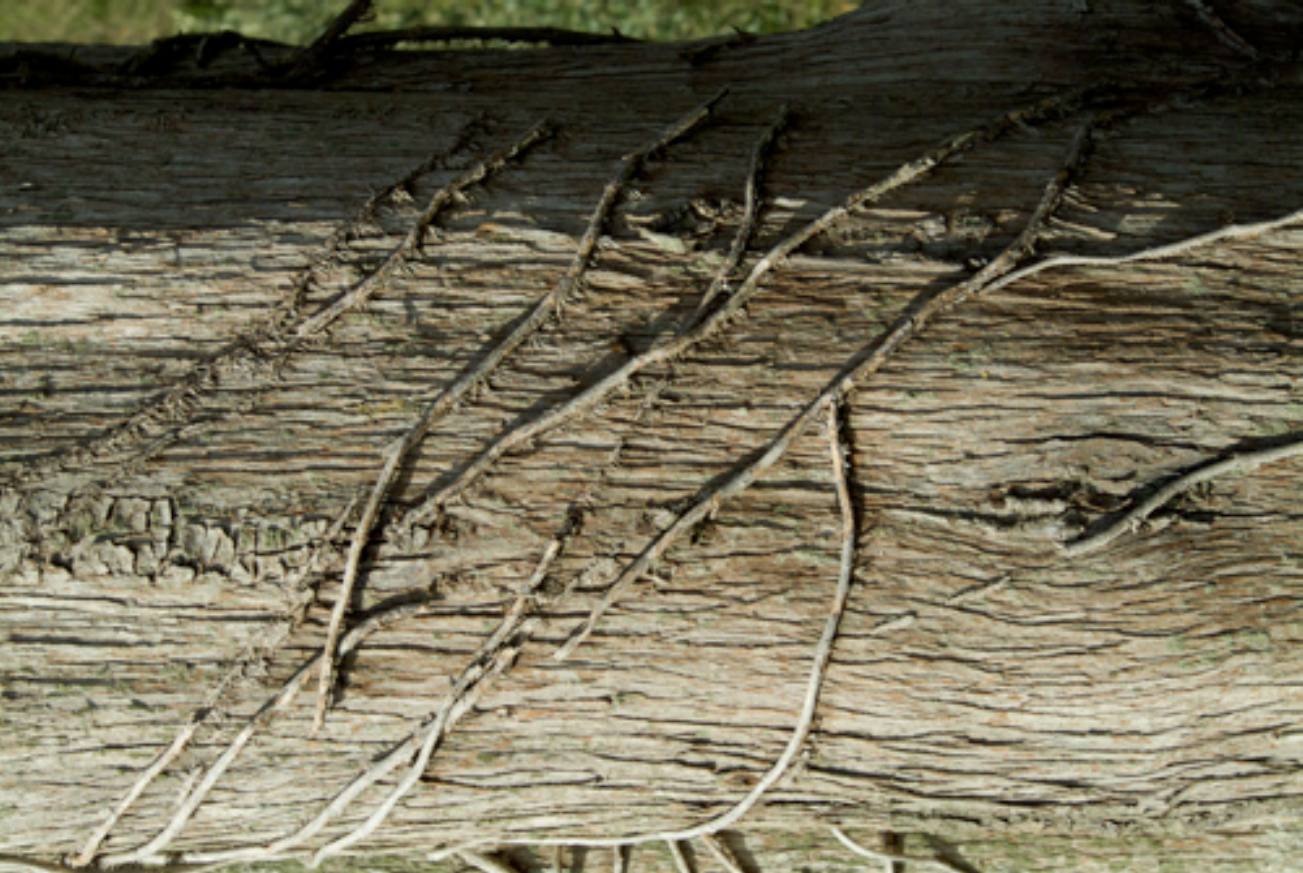}
\label{subf:ground-truth}}
\caption{Visual comparison between models w/o and w/ CEM.}
\label{f:visual}
\end{figure}

\subsection{Experimental Setup}
\label{subs:experimental}

We evaluated our method, CONE, by using two benchmark datasets, MIT~\cite{BychkovskyPCD11} and LSRW~\cite{HaiXYHZLH21}, reorganized by Ma et al.~\cite{MaMLFL22}. In this reorganized configuration, both datasets contain 500 training images. The number of test images is 100 and 50, respectively. We adopted two of the most widely used full-reference evaluation metrics, PSNR and SSIM, for performance evaluation.

For training, we used Adam~\cite{KingmaB14} as the optimizer with a weight decay of 3e-4. The mini-batch size was set to one. The learning rate of the IEM was set to 1e-4, while that of the CEM was initialized as 1e-5 and decayed by 0.1 every 100 epochs. The number of training epochs was 500 for MIT and 200 for LSRW. Our codes have been made publicly available on GitHub~\cite{github}.

The values of both $\omega_{f}$ and $\omega_{c}$ were set empirically in our experiments. All the other parameter values in Section~\ref{subs:loss} were inherited from ZeroDCE++~\cite{LiGL22}.

\subsection{Results}
\label{subs:results}

In this section, we explore the efficacy of CEM, and compare CONE with state-of-the-art methods on low-light image enhancement.

\smallskip\noindent
\textbf{Effectiveness of CEM.} Table~\ref{t:cem} shows the performance of the models without and with CEM. In this table, all models use both $\L^{(\t)}$ and $\L^{(\y)}$ as loss functions. As two baselines, `IEM' uses Eq.~\eqref{e:reflectance} to calculate $\y$, and `IEM+CEM' w/o `$\star$' fixes $\vartheta$ to the same values used in LECARM~\cite{RenYLL19}. The proposed method is denoted as `$\star$ IEM+CEM'.

From Table~\ref{t:cem}, we can see that our method could outperform the baseline `IEM' on both datasets if a proper comparametric equation was chosen. When sigmoid correction was used, CONE actually outperformed the baseline in all cases. Meanwhile, using the same IEM and CEM as CONE, but with fixed parameters $\vartheta$, the performance deteriorated significantly. This indicates that a simple combination of an illumination estimation-centric method and a comparametric equation does not necessarily improve deep image enhancement. Rather, it is important to learn CEM from the data using proper loss functions. In the following experiments, the sigmoid correction is used for MIT and the BetaGamma correction for LSRW.

Fig.~\ref{f:visual} shows the enhanced images and the error maps obtained with the baseline and CONE. From the error maps (Figs.~\ref{subf:baseline-diff} and \ref{subf:visual-cone-diff}), we can see that, thanks to the parameterization of the CRF (and the comparametric equation), CONE ensures greater function fitting capability and lower reconstruction error than the baseline, even when optimized with the same loss functions.

\smallskip\noindent
\textbf{Comparison to State of the Art.} Concentrating on low-light image enhancement, we compared CONE with three advanced handcrafted methods (LECARM~\cite{RenYLL19}, Hao's method~\cite{HaoHGXW20}, and STAR~\cite{XuHRLZYWS20}), three supervised learning methods (RetinexNet~\cite{WeiWY018}, Xu's method~\cite{XuYYL20}, and KinD++~\cite{ZhangGMLZ21}), as well as four unsupervised learning-based methods (Zhang's method~\cite{ZhangDZW20}, RUAS~\cite{Liu0Z0L21}, ZeroDCE++~\cite{LiGL22}, SCI~\cite{MaMLFL22}). As the CRF for LECARM, the sigmoid model and the BetaGamma model were used for MIT and LSRW, respectively. In KinD++, the illumination adjustment ratio was set to 2.5 for MIT and 5.0 for LSRW.

\begin{table}[t]
\centering
\caption{Comparison between state-of-the-art methods and CONE. Best performance is shown in bold.}
\label{t:sota}
\small
\begin{tabular*}{\linewidth}{@{\extracolsep{\fill}}lcccc}
\toprule
& \multicolumn{2}{c}{\textbf{MIT~\cite{BychkovskyPCD11}}} & \multicolumn{2}{c}{\textbf{LSRW~\cite{HaiXYHZLH21}}} \\
\cmidrule{2-3}
\cmidrule{4-5}
\textbf{Method} & \textbf{PSNR$\uparrow$} & \textbf{SSIM$\uparrow$} & \textbf{PSNR$\uparrow$} & \textbf{SSIM$\uparrow$} \\
\midrule
LECARM~\cite{RenYLL19} & 18.75 & 0.832 & 17.18 & 0.468 \\
Hao et al.~\cite{HaoHGXW20} & 17.62 & 0.782 & 14.71 & 0.486 \\
STAR~\cite{XuHRLZYWS20} & 16.26 & 0.698 & 14.62 & 0.474 \\
\midrule
RetinexNet~\cite{WeiWY018} & 12.84 & 0.660 & 15.48 & 0.347 \\
Xu et al.~\cite{XuYYL20} & 15.28 & 0.676 & 17.01 & \textbf{0.519} \\
KinD++~\cite{ZhangGMLZ21} & 17.35 & 0.797 & 16.17 & 0.417 \\
\midrule
Zhang et al.~\cite{ZhangDZW20} & 10.37 & 0.634 & 16.14 & 0.462 \\
RUAS~\cite{Liu0Z0L21} & 18.76 & 0.839 & 14.27 & 0.460 \\
ZeroDCE++~\cite{LiGL22} & 17.75 & 0.795 & 16.28 & 0.453 \\
SCI~\cite{MaMLFL22} & 20.84 & 0.850 & 15.17 & 0.418 \\
\midrule
$\star$ CONE & \textbf{21.19} & \textbf{0.853} & \textbf{17.39} & 0.460 \\
\bottomrule
\end{tabular*}
\end{table}

The results are compared in Table~\ref{t:sota}. Generally, CONE achieved the best performance in all metrics and on all datasets among the unsupervised learning methods. SCI and RUAS took second and third places overall. The performance of CONE is also comparable with that of the supervised learning methods, where Xu's method showed the highest SSIM among all the compared methods. As for the handcrafted methods, LECARM boasted a high level of competitiveness but could not pull up to CONE due to its lower task adaptability. In this study, we adopted SCI as the backbone network because of its great efficiency. However, this network does not incorporate denoising functionality, rendering CONE prone to amplifying compression artifacts and noises hidden in dark areas. This may explain the relatively lower SSIM of CONE on the LSRW dataset.

\begin{table}[t]
\centering
\caption{No. of parameters and inference complexity ($600\times400$). Best performance is shown in bold. \sdagger Data from Ma et al.~\cite{MaMLFL22}.}
\label{t:complexity}
\small
\begin{tabular*}{\linewidth}{@{\extracolsep{\fill}}lrr}
\toprule
\textbf{Method} & \textbf{Params (K)$\downarrow$} & \textbf{FLOPs (M)$\downarrow$} \\
\midrule
RetinexNet~\cite{WeiWY018}\sdagger & 838.3 & 136,015.1 \\
Xu et al.~\cite{XuYYL20} & 8,621.0 & 187,391.5 \\
KinD++~\cite{ZhangGMLZ21}\sdagger & 8,540.2 & 29,130.3 \\
\midrule
Zhang et al.~\cite{ZhangDZW20}\sdagger & 682.4 & 34,607.0 \\
RUAS~\cite{Liu0Z0L21} & 3.4 & 870.5 \\
ZeroDCE++~\cite{LiGL22}\sdagger & 78.9 & 5,211.2 \\
SCI~\cite{MaMLFL22} & \textbf{0.3} & \textbf{63.4} \\
\midrule
$\star$ CONE & \textbf{0.3} & \textbf{63.4} \\
\bottomrule
\end{tabular*}
\end{table}

\smallskip\noindent
\textbf{Inference Complexity.} We also investigated the memory usage and computation efficiency of our method. Table~\ref{t:complexity} compares the model size (number of parameters) and FLOPs required by state-of-the-art methods and our method during inference. For ease of comparisons, we computed the FLOPs by assuming the size of the test image to be $600\times400$, which is in accordance with the common configuration of RUAS~\cite{Liu0Z0L21} and SCI~\cite{MaMLFL22}.

In CONE, the CEM is built on top of the SCI backbone network and requires only two additional parameters $a$ and $b$. Therefore, the number of parameters required in CONE is almost the same as in SCI. Similarly, the FLOPs increase only negligibly because the comparametric equations in Table~\ref{t:comparametric} do not require complex convolution operations, but only pixel-wise elementary arithmetic and exponential operations. Thanks to the superior inference efficiency of SCI, CONE ranked high in terms of efficiency among the state-of-the-art methods in Table~\ref{t:complexity}.

\section{Conclusion}
\label{s:conclusion}

In this study, we addressed the problem of low-light image enhancement with the learning of DNNs. We proposed a novel method called CONE, which incorporates an intermediate CEM in an illumination estimation-centric neural network, to associate the illumination map $\t$ with the desired enhancement result $\y$ in a parameterized manner. Compared with previous studies, our method can approximate more flexible conversions (than the conventional Retinex model) between differently exposed photographs, and discovers a greater balance between enhancement flexibility and efficiency. Our method achieved PSNRs of 21.19 and 17.39 on MIT and LSRW datasets, respectively, updating state-of-the-art deep image enhancement. In the future, we shall investigate the performance of CONE in assisting downstream machine vision tasks, e.g., nighttime object detection and semantic segmentation, to increase the scale of our experiments and to further prove the greater flexibility of CONE. Ali and Mann~\cite{AliM12} showed that comparametric equations can be employed for HDR imaging, so our method can be extended for HDR reconstruction, but we leave it as a future direction.



\bibliographystyle{IEEEbib}
\bibliography{refs}
\end{document}